\documentclass{article}
\usepackage{spconf,amsmath,graphicx}
\usepackage{amsfonts}
\usepackage{url}
\usepackage{enumitem}
\setlist{nosep, leftmargin=14pt}

\usepackage{mwe} 
\usepackage{indentfirst}            
\usepackage{placeins}      
\usepackage{booktabs}
\usepackage{multirow}
 
\usepackage{caption}
\usepackage{subcaption} 
\usepackage{float}      
\usepackage{placeins}   

\usepackage{booktabs,caption}

\usepackage[letterpaper,
  left=1.03in, right=1.03in,   
  top=0.80in, bottom=1.00in,
  includeheadfoot, heightrounded
]{geometry}

\title{LPD: Learnable Prototypes with Diversity Regularization for Weakly Supervised Histopathology Segmentation}

\name{
\parbox{\textwidth}{\centering
Khang Le$^{\ast 2,4}$, Anh Mai Vu$^{\ast 1}$, Thi Kim Trang Vo$^{3,4}$,  Ha Thach$^{5}$, Ngoc Bui Lam Quang$^{6}$,\\
Thanh-Huy Nguyen$^{7}$, Minh H. N. Le$^{8}$, Zhu Han$^{1}$, Chandra Mohan$^{1}$, Hien Van Nguyen$^{1}$
}\thanks{$^{\ast}$Equal contribution.}
}
\address{
\parbox{\textwidth}{\centering
$^{1}$University of Houston , 
$^{2}$Ho Chi Minh City University of Technology \\ 
$^{3}$University of Information Technology, $^{4}$Vietnam National University Ho Chi Minh City,  $^{5}$University of Technology Sydney,  $^{6}$The University of Danang - VNUK Institute for Research and Executive Education, $^{7}$Carnegie Mellon University\\
$^{8}$Montefiore Medical Center, Albert Einstein College of Medicine 
}
}
\begin{document}
%
\maketitle
\begin{abstract}


Weakly supervised semantic segmentation (WSSS) in histopathology reduces pixel-level labeling by learning from image-level labels, but it is hindered by inter-class homogeneity, intra-class heterogeneity, and CAM-induced region shrinkage (global pooling–based class activation maps whose activations highlight only the most distinctive areas and miss nearby class regions). Recent works address these challenges by constructing a clustering prototype bank and then refining masks in a separate stage; however, such two-stage pipelines are costly, sensitive to hyperparameters, and decouple prototype discovery from segmentation learning, limiting their effectiveness and efficiency. We propose a cluster-free, one-stage learnable-prototype framework with diversity regularization to enhance morphological intra-class heterogeneity coverage. Our approach achieves state-of-the-art (SOTA) performance on BCSS-WSSS, outperforming prior methods in mIoU and mDice.  Qualitative segmentation maps show sharper boundaries and fewer mislabels, and activation heatmaps further reveal that, compared with clustering-based prototypes, our learnable prototypes cover more diverse and complementary regions within each class, providing consistent qualitative evidence for their effectiveness. Code is available at \url{https://github.com/tom1209-netizen/LPD}. 

\end{abstract}
\section{Introduction} \label{sec:intro}


Weakly supervised semantic segmentation (WSSS) in histopathological images has emerged as an effective paradigm to reduce reliance on costly pixel-level annotations by leveraging image-level labels, bounding boxes, or point annotations, thus alleviating the need for domain-expert pathologists, the tedious annotation process, and the large storage requirements of gigapixel whole-slide images~\cite{1}. There are three common approaches for the weakest image-level annotation: MIL-based~\cite{CLIM, Q-CLIM}, CAM-based~\cite{MLPS, TPRO}, and prototype-based~\cite{PBIP}. Most approaches first train a classification backbone to extract feature maps. Depending on the second step's design and optimization strategies, a different classification backbone is selected accordingly with commonly used ones, including ResNet~\cite{OEEN} , Vision Transformers (ViT)~\cite{ProtoSeg}, SegFormer~\cite{PBIP}, CLIP/MedCLIP encoders~\cite{TPRO}, each offering distinct advantages.


The second stage, pseudo-mask generation, varies widely across methods. MIL-based approaches~\cite{CLIM,Q-CLIM} treat each image as a bag of unlabeled patches and select informative instances via similarity to text or question–answer (QA) prompts for weak localization. In contrast, CAM-based methods rely on conventional classification class activation maps (CAMs), typically obtained via global pooling, to localize target regions. However, histopathological images exhibit intra-class heterogeneity and inter-class homogeneity~\cite{PBIP}, which complicates localization: such CAM pipelines often suffer \emph{region shrinkage}, where activations highlight only the most discriminative areas and miss the broader spatial extent of the class. To balance coverage and precision, CAM-based methods commonly add refinements such as multi-layer CAM fusion~\cite{MLPS}, high-confidence pixel selection~\cite{OEEN}, inter-layer CAM adjustment, or vision–language prompting~\cite{TPRO}.  
On the other hand, prototype-based methods form pseudo masks from prototype vectors that encode class or subclass patterns. Extraction strategies fall into two families: clustering-based and learnable. Clustering-based approaches (Proto2Seg, CPNet~\cite{CP_Net}, SiPE~\cite{SIPE}, PBIP~\cite{PBIP}) build semantic centers via K-means or affinity propagation, improving mask completeness but remaining tied to a pre-trained backbone. In contrast, learnable prototype approaches are primarily developed for interpretability or classification (ProtoPNet~\cite{ProtoPNet}, ProtoFormer~\cite{Protoformer}, ViLa-MIL~\cite{ViLa_MIL}, PIP-Net~\cite{PIP_Net}). ProtoSeg~\cite{ProtoSeg} brings learnable prototypes to WSSS but operates at the pixel level and is computationally heavy. Missing is a simple single-stage design that jointly learns prototypes with the segmenter, enforces prototype diversity, and avoids two-stage brittleness. We address this with a one-stage, cluster-free framework that integrates a learnable prototype module and diversity regularization directly into the backbone, enabling efficient training and broader, complementary pattern coverage. 

Our contributions in this work are: (1) We introduce an efficiency one-stage learnable-prototype framework for WSSS on histopathology,  eliminating multi-stage training and improving the efficiency of optimization
(2) We design a learnable prototype module with a diversity loss that encourages prototypes to attend to distinct tissue patterns. (3) Our method outperforms existing WSSS approaches on the BCSS-WSSS benchmark.

\section{Methodology}
\begin{figure*}[h]
\centering
\includegraphics[width=0.9\linewidth]{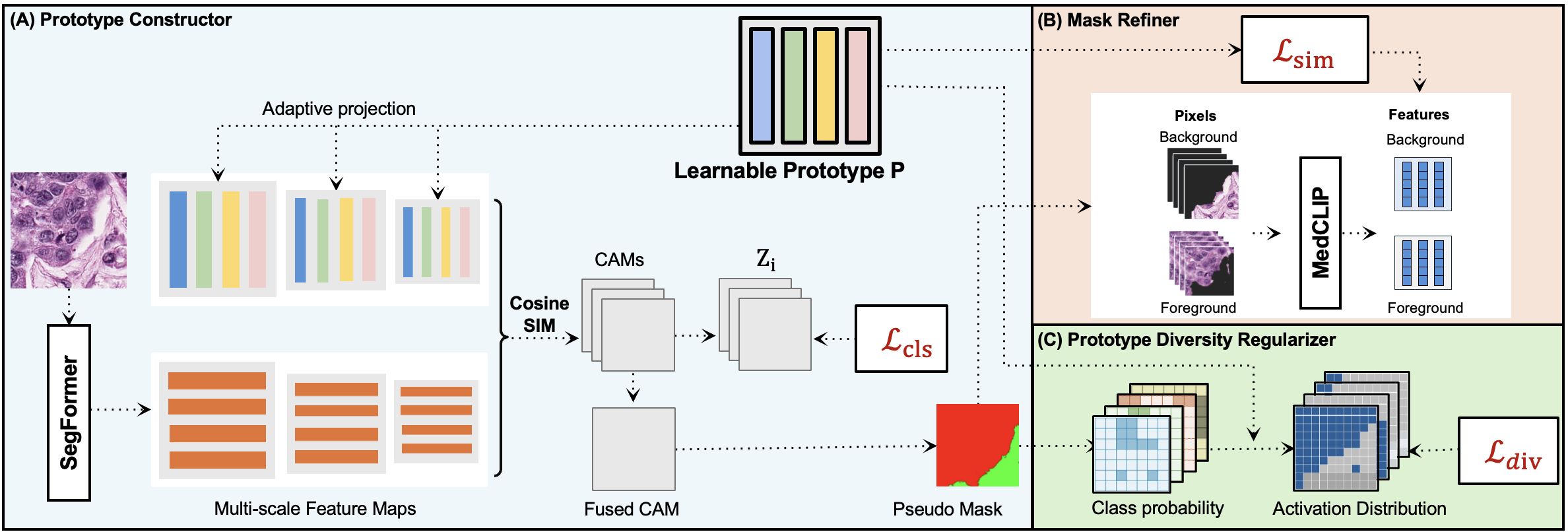}
\caption{
    \textbf{Overview of the LDP framework.}
    \emph{Prototype Constructor} (blue) produces learnable prototypes generating class CAMs.
    \emph{Mask Refiner} (orange, top) fuses CAMs into pseudo masks and performs FG/BG contrastive alignment.
    \emph{Prototype Diversity Regularizer} (green, bottom) encourages prototypes to capture distinct tissue patterns.
    }
\label{fig:model_overview}
\end{figure*}

\noindent \textbf{Overview.} In the following, we use CAMs to denote prototype-derived activation maps. Our method has three components: (1) a Prototype Constructor that builds class-wise learnable prototypes and produces CAMs; (2) a Mask Refiner that fuses these CAMs into pseudo masks and aligns foreground / background regions with the prototype space; and (3) a Prototype Diversity Regularizer that prevents prototype collapse. We build on PBIP \cite{PBIP} by replacing its clustering-based prototype bank with class-wise learnable prototypes trained end-to-end.
\subsection{Prototype Constructor} 
\noindent Given an input image $X$, we use a SegFormer encoder (MiT-B1) \cite{xie2021segformer} to extract multi-scale feature maps $F_i \in \mathbb{R}^{B\times d_i\times H_i\times W_i}$, $i\in\{1..4\}$. We instantiate $N_p{=}C\cdot k$ \emph{trainable} prototypes
$P\in\mathbb{R}^{(Ck)\times d_{\text{proto}}}$,
with $k$ prototypes per class, following prototype learning for interpretability~\cite{ProtoPNet,PIP_Net}.
Lightweight heads $\phi_i:\mathbb{R}^{d_{\text{proto}}}\to\mathbb{R}^{d_i}$ adapt the shared prototype space to each scale, producing
$P'_i=\phi_i(P)$.
For each scale, we compute per-pixel cosine similarity between feature vectors and prototypes to obtain prototype activation maps:
\begin{equation}
\textstyle
\begin{aligned}
S_i(x,j) &= \tau_i^{-1}\,
\left(\frac{f_i(x)}{\lVert f_i(x)\rVert_2}\right)^{\!\top}
\frac{p'_{i,j}}{\lVert p'_{i,j}\rVert_2} 
\end{aligned}
\label{eq:SiMi_single}
\end{equation}

Here, $M_i \in \mathbb{R}^{B\times (Ck)\times H_i\times W_i}$ denotes the stacked prototype activation maps. We obtain class-level activation maps (our prototype-based CAMs) by aggregating each class’s $k$ prototype maps using attention over its subclasses, yielding
$G_i \in \mathbb{R}^{B\times C\times H_i\times W_i}$.

Each $G_i$ is globally averaged to logits $z_i$ and supervised with image-level labels $y$:
\begin{equation}
\vcenter{\hbox{$\textstyle
\mathcal{L}_{\text{cls}}
=\sum_{i=1}^{4}\text{BCEWithLogits}\!\big(\text{GAP}(G_i),\,y\big)
$}}
\end{equation}

\subsection{Mask Refiner}
\noindent \textbf{Prototype activation fusion and pseudo masks.}
Following PBIP, we first fuse the class-level prototype-based CAMs $G_i$ from all stages into a single fused CAM,
\(
\text{cam} = \sum_i w_i\,G_i.
\)
We then generate pseudo masks by dynamic thresholding
\(
t = \alpha \max_x \text{cam}(x),
\)
and define foreground and background indicator functions as
\(\mathbb{1}_{\text{fg}}(x) = \mathbf{1}\{\text{cam}(x) \ge t\}, \quad
\mathbb{1}_{\text{bg}}(x) = 1 - \mathbb{1}_{\text{fg}}(x),
\)
which is standard in WSSS~\cite{ReCAM}.

\noindent\textbf{Foreground (FG)/background (BG) contrastive alignment.}
We crop masked FG and BG regions and encode their patch features using a frozen (Med)CLIP image encoder (as in PBIP)~\cite{PBIP,CLIP}, obtaining region embeddings $f^{\text{fg}}$ and $f^{\text{bg}}$. These embeddings are then projected into the prototype space by $\psi$. We use InfoNCE-style contrastive losses: FG features are pulled toward prototypes of the same class and pushed away from prototypes of other classes, while BG features are attracted to background prototypes and repelled from foreground prototypes. The refiner loss is
\(
\mathcal{L}_{\text{sim}} = \ell_{\text{fg}} + \ell_{\text{bg}},
\)
with shared temperatures across views and hard negatives sampled from other classes.

\subsection{Prototype Diversity Regularizer (PDR)} 
To prevent multiple prototypes within the same class from collapsing to similar regions, we introduce a PDR that explicitly discourages redundancy among intra-class prototypes~\cite{ProtoSeg,ContextProto}. Consider a class $c\in\{1,\dots,C\}$ associated with $k$ learnable prototypes $\{p_{c,u}\}_{u=1}^{k}$, where $u$ and $v$ index prototypes within the same class ($u,v\in\{1,\dots,k\}$, $u\neq v$).  
Let $\Omega_c$ denote the set of spatial locations (pixels or patches) predicted as class $c$ by the fused prototype-based CAM.  
For each location $x\in\Omega_c$, let $f(x)\in\mathbb{R}^{d}$ represent the feature vector extracted from the encoder stage used for diversity computation, and define its normalized form as $\hat f(x)=f(x)/\|f(x)\|_2$.  
Each $p_{c,u}$ is normalized to $\hat p_{c,u}=p_{c,u}/\|p_{c,u}\|_2$ after projection to the same feature dimension as $f$.  
The inner product $\langle \hat f(x), \hat p_{c,u} \rangle$ corresponds to the cosine similarity between feature and prototype embeddings. To capture each prototype's spacial focus,  we convert its similarity map into a categorical distribution over $\Omega_c$:
\begin{equation}
\textstyle
    v_{c,u}(x)
    = \mathrm{softmax}_{x\in\Omega_c}\!\big(\langle \hat f(x),\hat p_{c,u}\rangle\big)
\end{equation}

Each $v_{c,u}$ forms a probability distribution satisfying $\sum_{x\in\Omega_c}v_{c,u}(x)=1$, reflecting the spatial activation strength of prototype $u$ within class $c$. We treat $v_{c,u}$ and $v_{c,v}$ as discrete probability distributions supported on $\Omega_c$, and dissimilarity using Jeffrey’s divergence: 

\begin{equation}
\textstyle
    J(U,V)=\mathrm{KL}(U\|V)+\mathrm{KL}(V\|U)
\end{equation}

To encourage distinct spatial activations, we apply an exponentially decaying penalty based on this divergence.
    
Let $\mathcal{P}_c=\{(u,v)\mid 1\le u<v\le k\}$ denote the set of unordered prototype pairs, with cardinality $|\mathcal{P}_c|=k(k{-}1)/2$.  
The class-wise diversity regularization term is defined as
\begin{equation}
\textstyle
\mathcal{L}_{\text{div}}^{(c)}
=\frac{1}{|\mathcal{P}_c|}\sum_{(u,v)\in\mathcal{P}_c}\exp\!\big(-J(v_{c,u},v_{c,v})\big)
\end{equation}
which attains lower values when prototypes of the same class attend to complementary, non-overlapping subregions.  
Finally, we average across all classes with valid pseudo-regions 
\(
\textstyle
\mathcal{L}_{\text{div}}
=\frac{1}{|\mathcal{C}_\text{valid}|}
\sum_{c\in\mathcal{C}_\text{valid}}\mathcal{L}_{\text{div}}^{(c)}.
\)

The network is trained end-to-end with
\(
\mathcal{L}_{\text{total}}
=\mathcal{L}_{\text{cls}}
+\lambda_{\text{sim}}\mathcal{L}_{\text{sim}}
+\lambda_{\text{div}}\mathcal{L}_{\text{div}},
\) using a short classification warm-up before enabling $\mathcal{L}_{\text{sim}}$ and $\mathcal{L}_{\text{div}}$ for stability. We also adopt the post-processing method from the CRF paper (fully-connected dense CRF) at inference~\cite{CRF}.

\section{Experiments and Results} 




\noindent \textbf{Experimental Settings.} We evaluate on BCSS-WSSS (four classes: TUM, STR, LYM, NEC). We use SegFormer with an ImageNet-1K–pretrained Mix Transformer backbone. Optimization: AdamW (lr \(1\times10^{-5}\), weight decay 0.003), batch size 10, for 10 epochs. The best validation checkpoint is used for evaluation. We report mIoU and Dice to quantify segmentation result. 




\begin{table*}[h]
\centering
\renewcommand{\arraystretch}{1.15}
\setlength{\tabcolsep}{6pt}
\caption{Segmentation results on \textbf{BCSS-WSSS}}
\label{tab:bcss_wsss_compact_reformat}
\begin{tabular}{@{} l | r r | r r r r | r r r r @{}}
\toprule
\textbf{Method}
& \multicolumn{2}{c|}{\textbf{Metrics (\%)}} 
& \multicolumn{4}{c|}{\textbf{Per-class IoU (\%)}} 
& \multicolumn{4}{c}{\textbf{Per-class Dice (\%)}} \\
& \textbf{mIoU} & \textbf{mDice}
& \textbf{TUM} & \textbf{STR} & \textbf{LYM} & \textbf{NEC}
& \textbf{TUM} & \textbf{STR} & \textbf{LYM} & \textbf{NEC} \\
\midrule
\textbf{TPRO}
& 65.54 & 78.93
& 77.29 & 66.83 & 56.81 & 61.23
& 87.19 & 80.12 & 72.46 & 75.95 \\
\textbf{MLPS}
& 61.58 & 75.95
& 72.98 & 62.58 & 52.03 & 58.73
& 84.38 & 76.99 & 68.45 & 74.00 \\
\textbf{Proto2Seg}
& 57.42 & 72.24
& 63.25 & 58.28 & 53.27 & 54.89
& 77.49 & 73.64 & 67.78 & 70.08 \\
\textbf{PBIP}
& 69.42 & 81.84
& 77.92 & 64.68 & 65.40 & 69.69
& 87.59 & 78.56 & 79.08 & 82.14 \\
\bottomrule 
\textbf{OURS (3 prot.)}
& 70.83 & 82.77
& 81.97 & 68.30 & 65.85 & 67.22
& 90.09 & 81.16 & 79.41 & 80.40 \\
\textbf{OURS (10 prot.)}
& \textbf{71.96} & \textbf{83.56}
& \textbf{82.25} & 68.28 & \textbf{68.82} & 68.49
& \textbf{90.26} & 81.15 & \textbf{81.53} & 81.30 \\ 
\bottomrule
\end{tabular}
\end{table*}

\begin{table*}[h]
\centering
\renewcommand{\arraystretch}{1.15}
\setlength{\tabcolsep}{6pt}
\caption{Ablation on the number of prototypes and Diversity Parameter for \textbf{LDP} on \textbf{BCSS-WSSS}}
\label{tab:bcss_wsss_ours_div}
\begin{tabular}{@{} p{8mm} p{18mm} | r r | r r r r | r r r r @{}}
\toprule
\multicolumn{1}{c}{\# Prot.} & \multicolumn{1}{c|}{Div. Param}
& \multicolumn{2}{c|}{Metrics (\%)} 
& \multicolumn{4}{c|}{Per-class IoU (\%)} 
& \multicolumn{4}{c}{Per-class Dice (\%)} \\
&
& mIoU & mDice
& TUM & STR & LYM & NEC
& TUM & STR & LYM & NEC \\
\midrule
\makebox[\linewidth][c]{3} & \makebox[\linewidth][c]{0.00}
& 70.54 & 82.58
& 81.40 & 66.92 & 66.74 & 67.12
& 89.74 & 80.18 & 80.05 & 80.32 \\ 
\makebox[\linewidth][c]{3} & \makebox[\linewidth][c]{0.25}
& 70.28 & 82.36
& 82.10 & 68.07 & 66.56 & 64.40
& 90.17 & 81.00 & 79.93 & 78.35 \\
\makebox[\linewidth][c]{3} & \makebox[\linewidth][c]{0.50}
& \textbf{70.83} & \textbf{82.77}
& 81.97 & 68.30 & 65.85 & 67.22
& 90.09 & 81.16 & 79.41 & 80.40 \\
\makebox[\linewidth][c]{3} & \makebox[\linewidth][c]{0.75}
& 69.30 & 81.67
& 80.76 & 67.09 & 61.68 & 67.67
& 89.36 & 80.31 & 76.30 & 80.72 \\
\midrule
\makebox[\linewidth][c]{10} & \makebox[\linewidth][c]{0.00}
& 70.79 & 82.74
& 81.70 & 67.27 & 65.67 & 68.51
& 89.93 & 80.43 & 79.28 & 81.31 \\ 
\makebox[\linewidth][c]{10} & \makebox[\linewidth][c]{0.25}
& 71.83 & 83.47
& 81.88 & 68.07 & 67.85 & 69.50
& 90.04 & 81.00 & 80.84 & 82.01 \\
\makebox[\linewidth][c]{10} & \makebox[\linewidth][c]{0.50}
& \textbf{71.96} & \textbf{83.56}
& 82.25 & 68.28 & 68.82 & 68.49
& 90.26 & 81.15 & 81.53 & 81.30 \\
\makebox[\linewidth][c]{10} & \makebox[\linewidth][c]{0.75}
& 71.34 & 83.13
& 81.90 & 68.18 & 69.36 & 65.93
& 90.05 & 81.08 & 81.91 & 79.47 \\
\bottomrule
\end{tabular}\label{Table:abb}
\end{table*}
 






 \begin{figure}[h]
  \centering
  \includegraphics[width=0.60\linewidth]{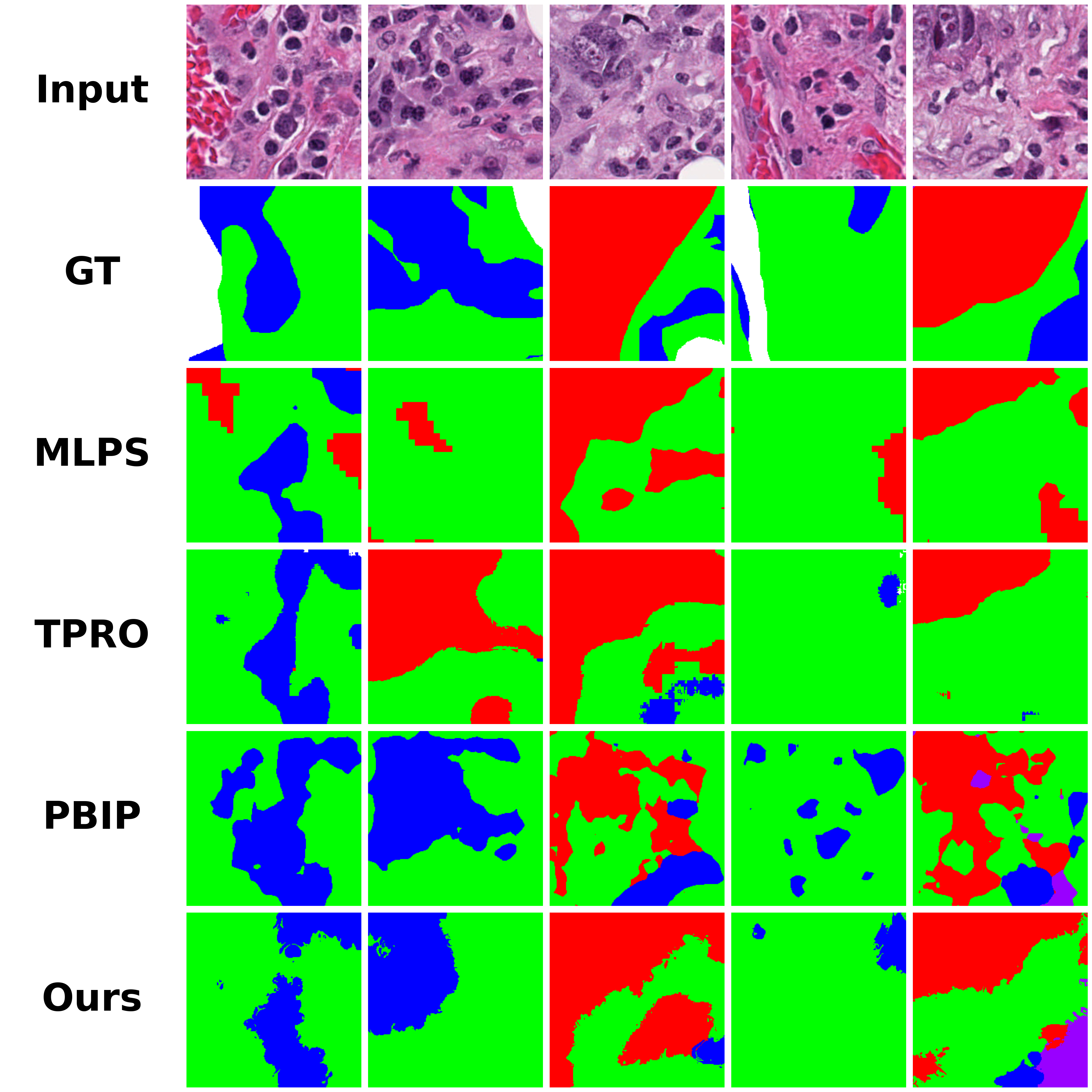}
  \caption{Visualization of segmentation masks of different model on the BCSS-WSSS dataset}
  \label{fig:overview}
\end{figure}    

\begin{figure}[h]
\centering
\includegraphics[width=0.8\linewidth]{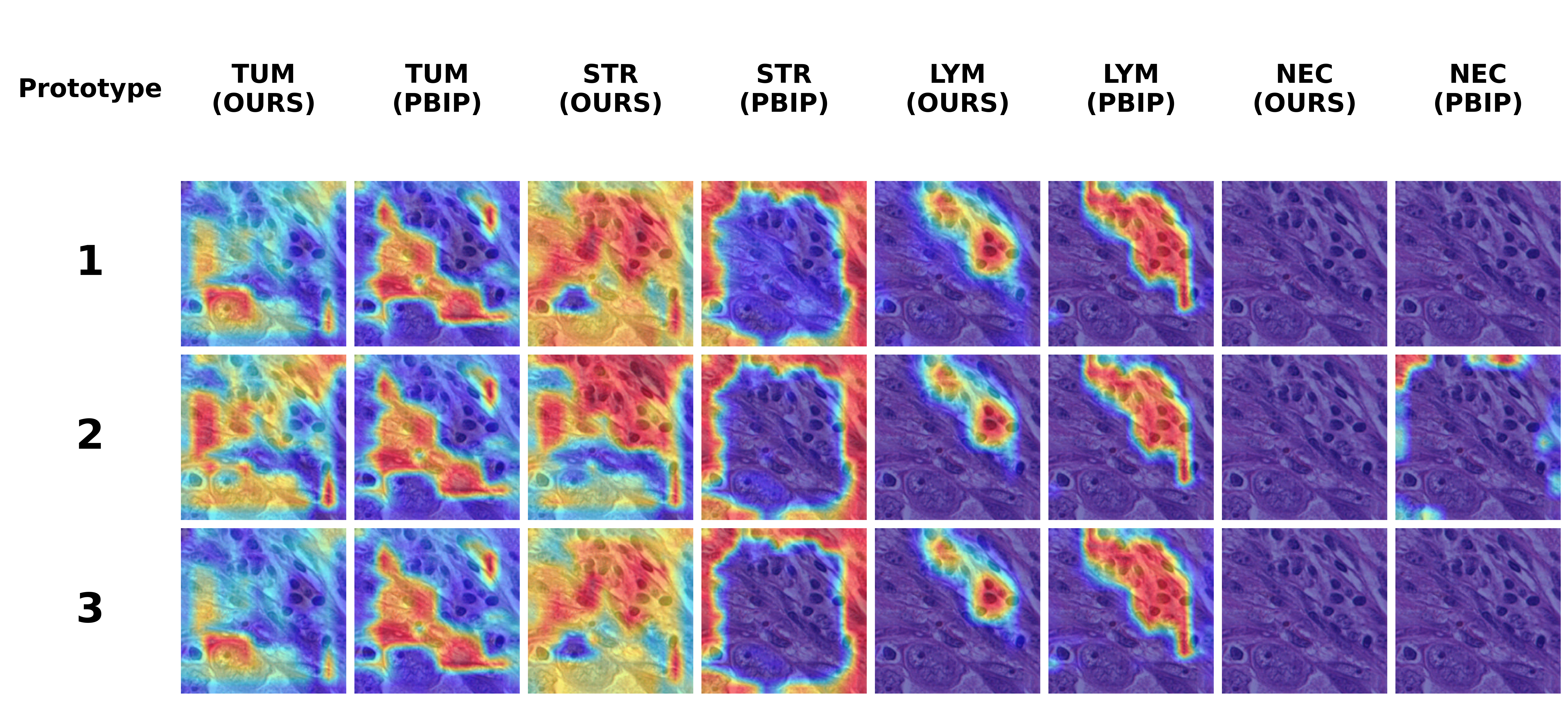}
\caption{\textbf{Activation heatmaps across segmentation classes} for clustering-based and LDP's prototype on BCSS-WSSS, each configured with 3 prototypes}
\label{fig:qual_bcss}
\end{figure}   
 

\noindent \textbf{Quantitative Results.} Our framework sets a new state of the art on BCSS-WSSS, outperforming existing weakly supervised semantic segmentation methods. The best configuration ($k{=}10$ prototypes) achieves 71.96\% mIoU and 83.56\% mDice, surpassing the previous SOTA PBIP (69.42\% mIoU, 81.84\% mDice) by 2.54 and 1.72 percentage points, respectively. Using 10 instead of 3 prototypes further boosts performance (Table~\ref{tab:bcss_wsss_compact_reformat}), consistent with richer prototype coverage.
 

  
\noindent \textbf{Qualitative Results.} We present segmentation masks in Figure~\ref{fig:overview}. Our method yields sharper boundaries, fewer false positives, and better class consistency (notably at tumor–stroma interfaces and for small lymphocytic regions), aligning with the quantitative results in Table ~\ref{tab:bcss_wsss_compact_reformat}. To study how prototype design affects localization, Figures ~\ref{fig:qual_bcss} compares activation heatmaps for clustering-based prototypes and our diversity-regularized learnable prototypes, both with three prototypes per class. PBIP prototypes are subclass means (K=3) that capture dominant patterns, whereas for the tumor (TUM) and stroma (STR) classes our learnable prototypes attend to complementary subregions and distinct morphological variants, hypothetically revealing more unique and fine-grained regions than subclass-mean prototypes.
 



\noindent  \textbf{Ablation on Prototype Count and Diversity Regularization.} As shown in Table~\ref{Table:abb}, our method with $k{=}10$ prototypes but no diversity loss achieves 70.79\% mIoU. Increasing the $k$ from $k{=}3$ (70.54\% mIoU) to $k{=}10$ gives only a marginal gain, indicating redundant prototypes and prototype collapse. When we add the diversity regularizer ($\mathcal{L}_{\text{div}}$)  at $k{=}10$, prototypes are encouraged to specialize and capture distinct tissue patterns, yielding a 1.17-point improvement (from 70.79\% to 71.96\% mIoU) and our final SOTA result. This shows that $\mathcal{L}_{\text{div}}$ is crucial for scaling prototype capacity. 

\section{Conclusion}  
We propose a cluster-free, one-stage learnable-prototype framework for WSSS that removes multi-stage overhead and better handles intra-class heterogeneity. End-to-end training couples prototype discovery with segmentation, and a diversity regularizer prevents collapse by encouraging complementary tissue patterns. On BCSS-WSSS, our method achieves SOTA performance, showing that diversity-aware prototypes are an efficient solution.

\noindent \textbf{Acknowledgments.} This work was supported in part by the National Institutes of Health (NIH) under Grant 5R01DK134055-02. 
\label{sec:acknowledgments}




\bibliographystyle{IEEEbib}
\bibliography{refs}

\end{document}